\documentclass[sigconf]{acmart}

\usepackage{booktabs} 
\usepackage{todonotes}
\usepackage{placeins}
\usepackage{import}
\usepackage{url}
\usepackage{subfigure,wrapfig}
\usepackage{amssymb}
\usepackage{amsmath}
\usepackage{verbatim}
\usepackage{latexsym}
\usepackage{graphicx}
\usepackage{wrapfig}
\usepackage{caption}
\usepackage{xspace}       
\usepackage{hyperref}
\usepackage{balance}
\usepackage{multicol}
\usepackage{flushend}

\renewcommand\footnotetextcopyrightpermission[1]{} 
\usepackage{footmisc}

\begin{document}
\title{F1/10: An Open-Source Autonomous Cyber-Physical Platform}
\author{\normalsize{\textbf{Matthew O'Kelly*, Houssam Abbas**, Jack Harkins, Chris Kao, Yash Vardhan Pant \& Rahul Mangharam} \\
Department of Electrical and Systems Engineering,
University of Pennsylvania, USA \\
\{mokelly, harkj, chriskao, yashpant, rahulm\}@seas.upenn.edu\\

** School of Electrical Engineering and Computer Science, Oregon State University, USA\\
houssam.abbas@oregonstate.edu\\
\vspace{2mm}
\textbf{Varundev Suresh Babu*, Dipshil Agarwal \& Madhur Behl}\\
Department of Computer Science,
University of Virginia, USA \\
\{varundev, dpa4va, madhur.behl\}@virginia.edu\\
\vspace{2mm}
\textbf{Paolo Burgio \& Marko Bertogna}\\
Dept. of Physics, Informatics \& Mathematics,
University of Modena \& Reggio Emilia, Italy\\
\{marko.bertogna, paolo.burgio\}@unimore.it \\}
\vspace{2mm}
{* \small These authors contributed equally}

}
\renewcommand{\shortauthors}{M. O'Kelly et al.}

\begin{abstract}
In 2005 DARPA labeled the realization of viable autonomous vehicles (AVs) a \textit{grand challenge}; a short time later the idea became a \textit{moonshot} that could change the automotive industry. Today, the question of \textit{safety} stands between \textit{reality} and \textit{solved}. Given the right platform the CPS community is poised to offer unique insights.
However, testing the limits of safety and performance on real vehicles is costly and hazardous. The use of such vehicles is also outside the reach of most researchers and students.
In this paper, we present F1/10: an open-source, affordable, and high-performance 1/10 scale autonomous vehicle testbed.
The F1/10 testbed carries a full suite of sensors, perception, planning, control, and networking software stacks that are similar to full scale solutions. 
We demonstrate key examples of the research enabled by the F1/10 testbed, and how the platform can be used to augment research and education in autonomous systems, making autonomy more accessible. 
\end{abstract}

\maketitle

\section{Introduction}

Progress in cyber-physical systems (CPS) requires the availability of robust platforms on which researchers can conduct real-world experiments and testing.
Unfortunatley, a vast majority of CPS experiments are done in isolation - either completely in simulation, or on proprietary hardware designs.
In either case, researchers are limited by the inability to deploy their methodologies in realistic environments without solving a host of unrelated problems. In many cases, due to these challenges, the research becomes less reproducible. In contrast, open source tools, and platforms, which can be commonly used across different CPS disciplines and by multiple research groups can be a primary driver in enabling high-impact research and teaching.

This lack of commonly available CPS testbeds is especially significant in the rapidly growing field of connected, and autonomous vehicles (AVs).
Modern full-scale automotive platforms are some of the most complex cyber-physical systems ever designed. 
From real-time and embedded systems, to machine learning and AI, sensor networks, to predictive control, formal methods, security \& privacy, to infrastructure planning, and transportation - the design of trustworthy, safe AVs is a truly interdisciplinary endeavour that has captured the imagination of researchers in both academia and industry. 
Auto companies are joining with tech giants like Google, Uber, and prominent start-ups to develop next-generation autonomous vehicles that will alter our roads and lay the groundwork for future smart cities. 

Today, conducting research in autonomous systems and AVs requires building one's own
automotive testbed from scratch. Sometimes researchers must enter into restrictive agreements with automotive manufactures to obtain access to the documentation necessary to build such a testbed, thus preventing the release of their testbed software and hardware. 

This paper presents the F1/10 Autonomous Racing Cyber-Physical platform and summarizes the use of this testbed technology as the common denominator and key enabler to address the research and education needs of future autonomous systems and automotive Cyber-Physical Systems.
\begin{figure}
    \centering
    \includegraphics[width=0.8\columnwidth]{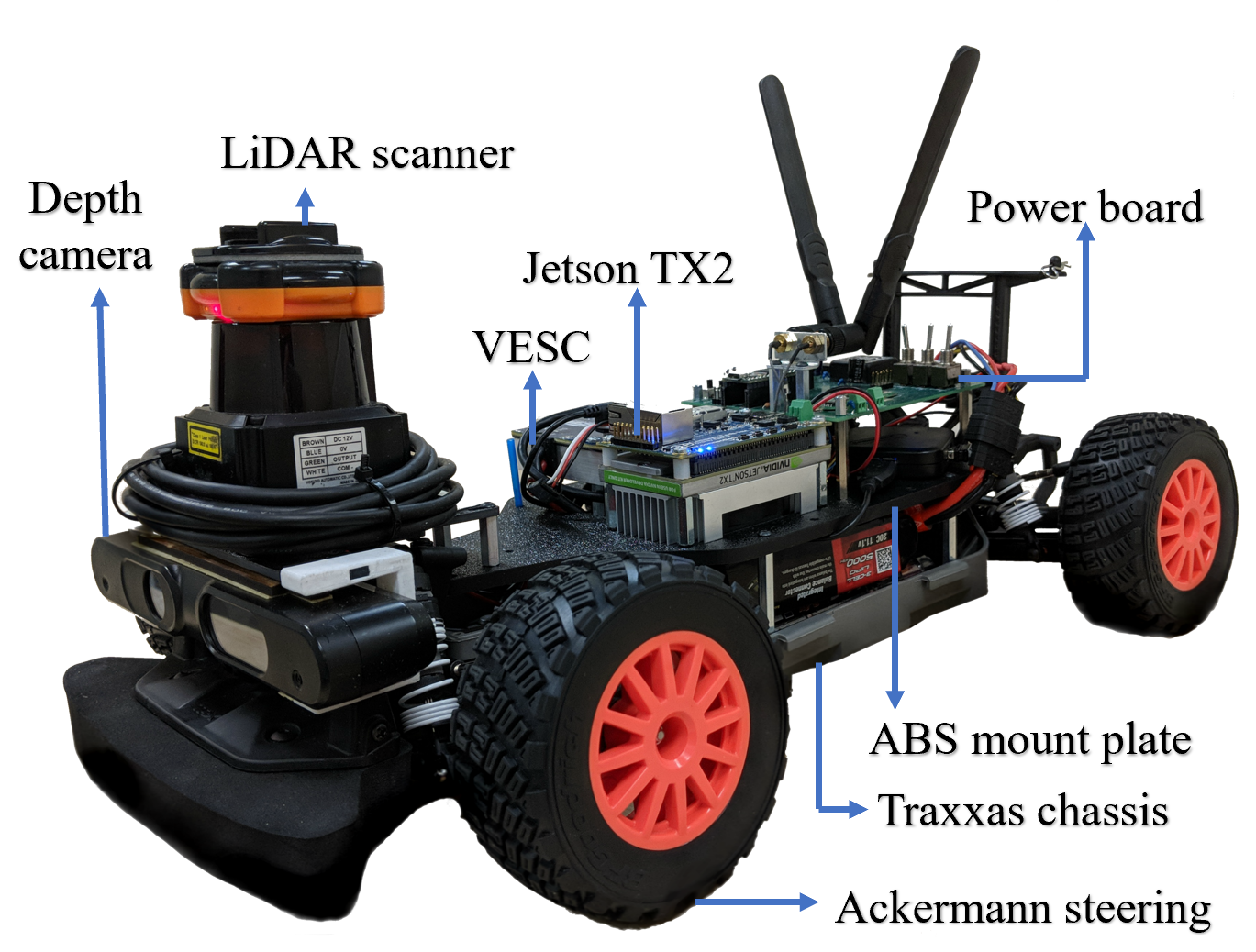}
    \caption{It takes only a couple of hours fully to assemble a F1/10 autonomous racecar, using detailed instructions available at \url{http://f1tenth.org/}}
\end{figure}
\begin{figure*}
    \centering
    \includegraphics[width=0.8\linewidth]{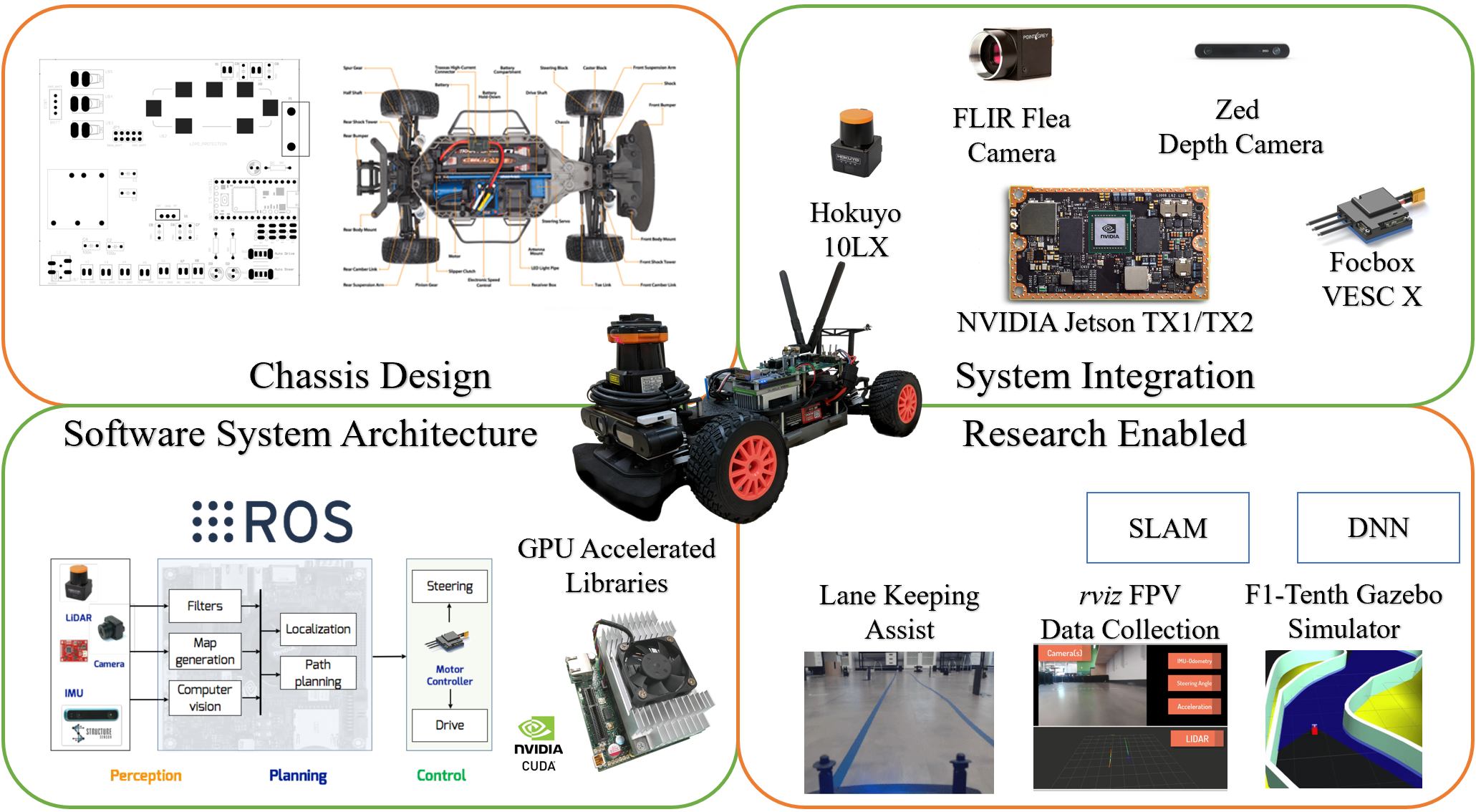}
    \caption{Overview of the F1/10 research instrument - Modular chassis and system design with detailed instructions; open-source software stack in ROS; and a wide variety of AV research enabled in the lab on a real testbed. }
    \label{fig:overview}
\end{figure*}
There are no affordable, open-source, and integrated autonomous vehicles test-beds available today that would fit in a typical indoor CPS lab.
Our goal is not to provide yet another isolated vehicle testbed. 
Instead, we aim to relieve researchers around the world of the requirement to set up their own facilities for research in autonomous vehicles. The F1/10 research instrument has the potential to build stronger networks of collaborative research.
Since the platform is 1/10 the the scale of a real vehicle we call it F1/10 (inspired from Formula 1 (F1)). 
Kick-started through a joint effort by University of Pennsylvania (USA), University of Virginia (USA), and UNIMORE (Italy), the F1/10 community is rapidly growing with about 20+ institutions utilizing the test-bed. 
 
F1/10 enables researchers and students to rapidly explore automotive cyber-physical systems by providing a platform for real-world experimentation. F1/10's biggest value is in taking care of the most tedious aspects of putting together an autonomous vehicle testbed so that the user can focus directly on the research and learning goals. 

While commercially available mobile platforms like TurtleBot2~\cite{turtlebot2}, and Jackal UGV~\cite{jackalugv} can be used as a research testbed, they lack realistic dynamics like Ackermann steering, and the ability to travel at high speeds - a characteristic which is essential for any autonomous vehicle testbed.
In contrast, the F1/10 platform is designed to address the issues of realistic vehicle dynamics, and drive-trains. 
We have designed the F1/10 platform using fully open-source and standardized systems that take advantage of ROS~\cite{quigley2009ros} and its associated libraries. 
On our website \url{http://f1tenth.org/}, detailed, and free instructions are available on how to build, and drive the platform. There is an active community of researchers who contribute to both the open-source hardware and software design. 

We present the following open-source capabilities of the F1/10 Autonomous Cyber-Physical Platform: (i) Open-source mechanical design (chassis, development circuit boards, programmable hardware) and open-source kits for assembling a 1/10-scale autonomous racing car. (ii) A suite of AV software libraries for perception, planning, control and coordinated autonomy research. (iii)  F1/10 simulator and virtual race track. (iv) Multiple annual autonomous racing competitions, hackathons, and high-school education programs. (v) Online course material and data sets.
This paper has the following research contributions:
\vspace{-1mm}
\begin{enumerate}
    \item The design and implementation of F1/10, an open-source autonomous testbed for research and education in autonomy,
    \item Modular hardware and software stacks that make the F1/10 testbed an accessible, AV vehicle research tool,
    \item More than a dozen representative examples of the types of research enabled by the F1/10 platform, particularly those that can be used to test AV algorithms and software pipelines with realistic dynamics on a physical and affordable testbed,
    \item A case study of going from 1/10 scale F1/10 cars to full scale autonomous vehicles,
    \item Overview of the widely successfully and exciting F1/10 Autonomous Racing Competitions being held at premier CPS and Embedded Systems venues over the last 3 years.
\end{enumerate}

\section{F1/10 Testbed}

The F1/10 platform is designed to meet the following requirements: (a) The platform must be able to capture the dynamics of a full scaled autonomous car;
(b) The platform's hardware and software stack must be modular so as to enable easy upgrades, maintenance and repairs;  and (c) The platform must be self-sustaining in terms of power, computation and sensors, i.e, it need not use any external localization (VICON cameras).

\subsection{System Architecture}
Figure~\ref{fig:overview} shows an overview of the F1/10 platform. 
The perception module interfaces and controls the various sensors including scanning LiDARs, monocular \& stereo cameras, inertial sensors, etc. The sensors provide the platform with the ability to navigate and localize in the operating environment.
The planning pipeline (in ROS) helps process the sensor data, and run mapping, and path planning algorithms to determine the trajectory of the car.
Finally, the control module determines the steering and acceleration commands to follow the trajectory in a robust manner. 

\subsection{F1/10 Build}
In this section we provide a brief description of how the F1/10 autonomous race is built. Detailed instructions and assembly videos can be found at \url{f1tenth.org}.

\noindent \textbf{Chassis:} The chassis consists of two parts. The bottom chassis is a 1/10 scale race car chassis available from Traxxas~\cite{traxxasref}. The top chassis is a custom laser-cut ABS plate that our team has developed and to which all the electronic components are attached. The CAD and laser cut files for the top plate are open-sourced. 
The Traxxas bottom chassis is no ordinary racing toy: it is a very realistic representation of a real car. It has 4-wheel drive and can reach a top speed of 40mph, which is extremely fast for a car this size. Tire designs replicate the racing rubber used on tarmac circuits. The turnbuckles have broad flats that make it easy to set toe-in and camber, just like in a real race car. The bottom chassis has a high RPM brush-less DC motor to provide the drive to all the wheels, an Electronic Speed Controller (ESC) to controls the main drive motor using pulse-width modulation (PWM), a servo motor for controlling the Ackermann steering, and a battery pack; which provides power to all these systems.
All the sensors and the on-board computer are powered by a separate power source (lithium-ion battery).
The F1/10 platform components are affordable and widely available across the world making it accessible for research groups at most institutions. 
These components are properly documented and supported by the manufacturer and the open-source community. 

\noindent \textbf{Sensors and Computation:}
The F1/10 platform uses an NVIDIA Jetson TX2~\cite{franklin2017nvidia} GPU computer. The Jetson is housed on a carrier board~\cite{orbittycarrier} to reduce the form factor and power consumption. 
The Jetson computer hosts the F1/10 software stack built on Robot Operating System (ROS). 
The entire software stack, compatible with the sensors listed below, is available as an image that can be flashed onto the Jetson, enabling a plug-and-play build. 
The default sensor configuration includes a monocular USB web cam, a ZED depth camera, Hokuyo 10LX scanning LiDAR, and a MPU-9050 inertial measurement unit (IMU). These sensors connect to the Jetson computer over a USB3 hub. Since the underpinnings of the software stack is in ROS, many other user preferred sensors can also be integrated/replaced.

\noindent \textbf{Power Board:} In order to enable high performance driving and computing the F1/10 platform utilizes Lithium Polymer batteries. The power board is used to provide a stable voltage source for the car and its peripherals since the battery voltage varies as the vehicle is operated. The power board also greatly simplifies wiring of peripherals such as the LIDAR and wifi antennas. Lastly the power board includes a Teensy MCU in order to provide a simple interface to sensors such as wheel encoders and add-ons such as RF receivers for long range remote control. 
\begin{figure*}[t]
    \centering
    \includegraphics[width=0.8\textwidth]{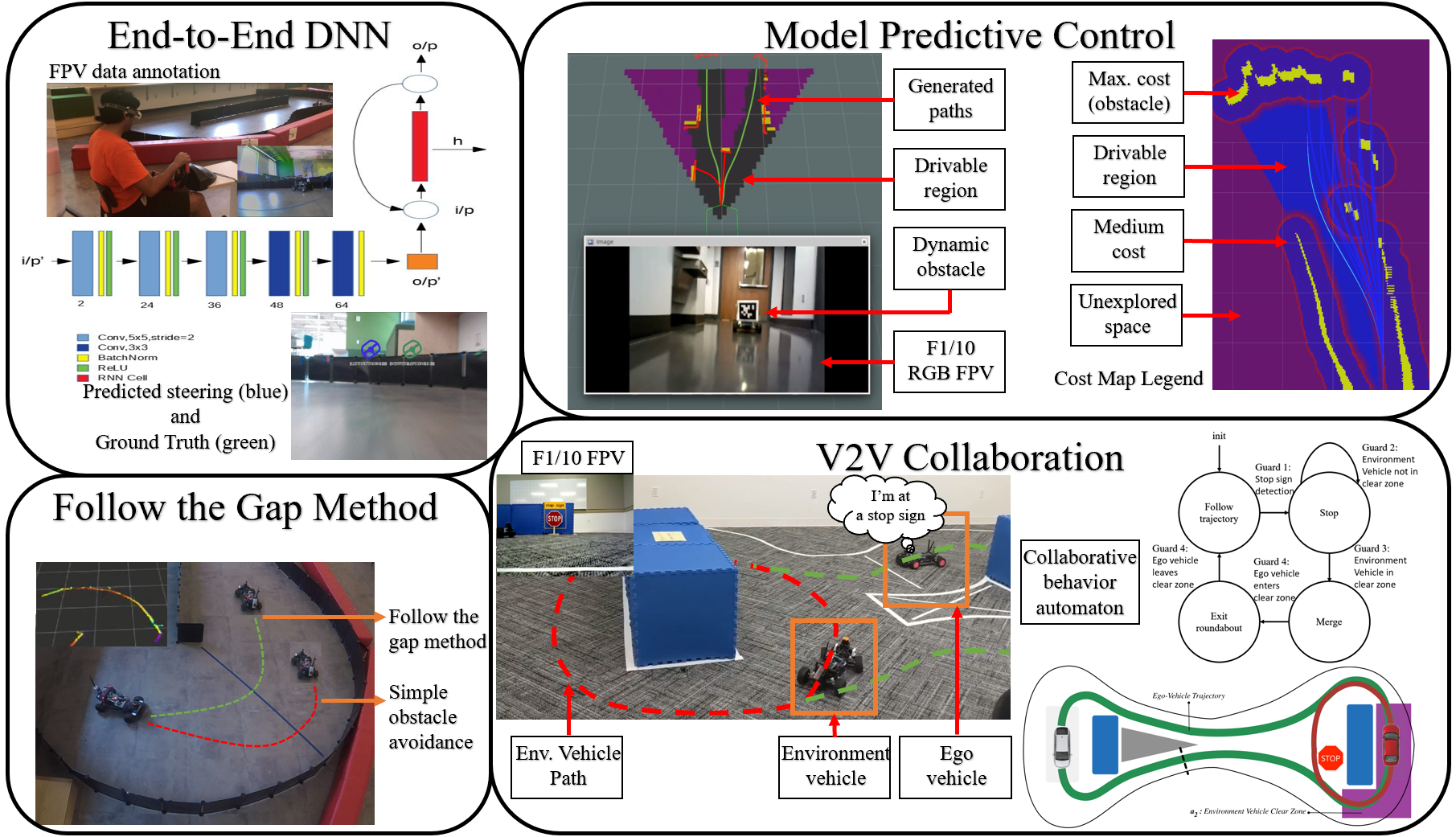}
    \caption{Planing and control research enabled by the F1/10 platform: \textit{(Bottom Left)} Reactive Obstacle Avoidance, \textit{(Top Left)} End-to-End Driving, \textit{(Top Right)} Model Predictive Control, \textit{(Bottom Right)} V2V Collaboration}
    \label{fig:planning-control}
\end{figure*}

\noindent \textbf{Odometry:}
Precise odometry is critical for path planing, mapping, and localization.
Odometry is provided by the on board VESC as an estimate of the steering angle and the position of the vehicle. 
The open-source F1/10 software stack includes the custom ROS nodes, and a configuration file required to interface with the VESC and obtain the odometry information.
\noindent \textbf{Communication architecture:}
The F1/10 testbed includes a wireless access point which is used to remotely connect (ssh) into the Jetson board. 
The software stack is configured to use \textit{ROS-Over-Network} used for both sending commands to the car and obtaining telemetry data from the car in real-time. In addition we have created software which supplies a socket which enables communication between multiple F1/10 vehicles operating under different ROS master nodes.

\section{Research: Planning and Control}
The decision making systems utilized on AVs have progressed significantly in recent years; however they still remain a key challenge in enabling AV deployment \cite{shalev2017formal}. While AVs today can perform well in simple scenarios such as highway driving; they often struggle in scenarios such as merges, pedestrian crossings, roundabouts, and unprotected left-turns. Conducting research in difficult scenarios using full-size vehicles is both expensive and risky. In this section we highlight how the F1/10 platform can enable research on algorithms for obstacle avoidance, end-to-end driving, model predictive control, and vehicle-to-vehicle communication. 


\subsection{Obstacle avoidance}
\label{subsec:ftg}
Obstacle avoidance and forward collision assist are essential to the operation of an autonomous vehicle. 
The AV is required to scan the environment for obstacles and safely navigate around them.  
For this reason, many researchers have 
developed interesting real-time approaches for avoiding unexpected static and dynamic obstacles~\cite{tallamraju2018decentralized, iacono2018path}. 
To showcase the capability of the F1/10 testbed, we implement one such algorithm known as \textit{Follow The Gap} (FTG) method~\cite{SEZER20121123}. 
The Follow the Gap method is based on the construction of a gap array around the vehicle and calculation of the best heading angle for moving the robot into the center of the maximum gap in front, while simultaneously considering its goal. 
These two objectives are considered simultaneously by using a fusing function. 
Figure~\ref{fig:planning-control}[Left] shows an overview and the constraints of FTG method. 
The three steps involved in FTG are:\\
(a) Calculating the gap array using vector field histogram, and finding the maximum gap in the LIDAR point cloud using an efficient sorting algorithm,\\
(b) Calculating the center of the largest gap, and\\
(c) Calculating the heading angle to the centre of the largest gap in reference to the orientation of the car, and generating a steering control value for the car.


\subsection{End-to-end driving}

Some recent research replaces the classic chain of perception, planning, and control with a neural network that directly maps sensor input to control output~\cite{DBLP:journals/corr/BojarskiTDFFGJM16, chi2017deep, eraqi2017end}, a methodology known as end-to-end driving.
Despite the early interest in end-to-end driving \cite{pomerleau1989alvinn}, most self-driving cars still use the perception-planning-control paradigm. 
This slow development can be explained by the challenges of verifying system performance; however, new approaches based on reinforcement learning are being actively developed \cite{kendall2018learning}.

The F1/10 testbed is a well suited candidate for experimentation with end-to-end driving pipelines, from data gathering and annotation, to inference, and in some cases even training.

\noindent \textbf{Data gathering and annotation for deep learning:}
As shown in Figure~\ref{fig:planning-control}[Right], we are able to integrate a First Person View (FPV) camera and headset with the F1/10 car. We are also able to drive the car manually with a USB steering wheel and pedals instead of the RC remote controller which comes with the Traxxas car. 
The setup consists of a Fat Shark FSV1204 - 700TVL CMOS Fixed Mount FPV Camera, 5.8GHz spiroNET Cloverleaf Antenna Set, 5.8GhZ ImmersionRC receiver, and Fat Shark FSV1076-02 Dominator HD3 Core Modular 3D FPV Goggles Headset.
The FPV setup easily enables teleoperation for the purposes of collecting data to train the end-to-end deep neural netowrks (DNNs). 
Each training data consists of an input, in this case an image from the front facing camera, and a label a vector containing the steering angle and requested acceleration. 
Practical issues arise due to the fact that the label measurements (50 Hz) must be synchronized with the acquired camera images (30 Hz). Included in this portion of the stack is a ROS node which aligns the measurements and the labels. As part of this research we are releasing over 40,000 labeled images collected from multiple builds at the University of Pennsylvania and the University of Virginia.

\noindent \textbf{End-to-End driving:}
Partly inspired by Pilotnet~\cite{DBLP:journals/corr/BojarskiTDFFGJM16} end-to-end work, we implemented a combination of a LSTM~\cite{NIPS2012_4824} and a Convolutional Neural Network(CNN)~\cite{Hochreiter:1997:LSM:1246443.1246450} cell.
These units are then used in the form of a recurrent neural network (RNN). 
This setup uses the benefits of LSTMs in maintaining temporal information (critical to driving) and utilizes the ability of CNN's to extract high level features from images. 

To evaluate the performance of the model we use the normalized root mean square error (NRMSE) metric between the ground truth steering value and the predicted value from the DNN. 
As can be seen in the point-of-view (PoV) image in Figure~\ref{fig:planning-control}[Left], our DNN is able to accurately predict the steering angle with an NRMSE of 0.14. 

\subsection{Global \& local approaches to path planning}

\label{sec:path_planning}
AVs operate in relatively structured environments. Most scenarios an AV might face feature some static structure. Often this is the road geometry, lane connectivity, locations of traffic signals, buildings, etc. Many AVs exploit the static nature of these elements to increase their robustness to sensing errors or uncertainty.
In the context of F1/10, it may be convenient to exploit some information known \textit{a priori} about the environment, such as the track layout and floor friction. 
These approaches are called \textit{static}, or \textit{global}, and they typically imply building  a map of the track, simulating the car in the map, and computing offline a suitable nominal path which the vehicle will attempt to follow. Valuable data related to friction and drift may also be collected to refine the vehicle dynamics model.
More refined models can be adopted off-line to compute optimal paths and target vehicle speeds, adopting more precise optimization routines that have a higher computational complexity to minimize the lap time.

Once the desired global path has been defined, the online planner must track it. To do that, there are two main activities must be accomplished on-line, namely \textit{localization} and \textit{vehicle dynamics control}.
Once the vehicle has been properly localized within a map, a local planner is adopted to send longitudinal and transversal control signals to follow the precomputed optimal path. As the local planner needs to run in real-time, simpler controllers are adopted to decrease the control latency as much as possible. Convenient online controllers include pure pursuit path geometric tracking \cite{coulter1992implementation}. The F1/10 software distribution includes an implementation of pure pursuit, nodes for creating and loading waypoints, and path visualization tools. For the interested reader we recommend this comprehensive survey of classical planning methods employed on AVs \cite{frazzoliMPsurvey16}.

\subsection{Model Predictive Control}
\label{sec:mpc}
While data annotation for training end-to-end networks is relatively easy, the performance of such methods is difficult to validate empirically \cite{shalev2016sample} especially relative to approaches which decompose functionality into interpret-able modules. 
In this section we outline both a local planner which utilizes a model predictive controller (MPC) and a learned approximation of the policy it generates detailing one way planning components can be replaced with efficient learned modules.

\noindent\textbf{Components:} The F1/10 platform includes a MPC written in C++ comprised of the vehicle dynamics model, an optimization routine which performs gradient descent on the spline parameters. Peripheral support nodes provide an interface to road center line information, a multi-threaded goal sampler, a 2D occupancy grid, and a trajectory evaluation module. Additionally, we include a CUDA implementation of a learned approximation of the MPC which utilizes the same interface as described above. 

\noindent\textbf{Cubic Spline Trajectory Generation:} One local planner available on the F1/10 vehicle utilizes the methods outlined in 
\cite{McNaughton2011} and \cite{Howard_2009_6434} and first described in \cite{nagy2001trajectory}. This approach is commonly known as \emph{state-lattice planning with cubic spline trajectory generation}.
Each execution of the planner requires the current state of the vehicle and a goal state. Planning occurs in a local coordinate frame. The vehicle state $x$ is defined in the local coordinate system, a subscript indicates a particular kind of state (i.e. a goal) In this implementation we define $x$ as: $\vec{x}={[s_x\ s_y\ v\ \Psi\ \kappa]}^T$, where $s_x$ and $s_y$ are the x and y positions of the center of mass, $v$ is the velocity, $\Psi$ is the heading angle, and $\kappa$ is the curvature. 

In this formulation, trajectories are limited to a specific class of parameterized curves known as \emph{cubic splines} which are dense in the robot workspace. We represent a cubic spline as a function of arc length such that the parameters $\vec{p} = [s\ a\ b\ c\ d]^T$ 
where $s$ is the total curve length and ($a,b,c,d)$ are equispaced knot points representing the curvature at a particular arc length. When these parameters are used to define the expression of $\kappa(s)$ which can be used to steer the vehicle directly. 
The local planner's objective is then to find a \emph{feasible trajectory} from the initial state defined by the tuple 
$\vec{x}$ to a goal pose $\vec{x}_{g}$.

We use a gradient descent algorithm and forward simulation models which limit the ego-vehicle curvature presented in \cite{Howard_2009_6434}. 
These methods ensure that the path generated is kinematically and dynamically feasible up to a specified velocity. 

 \noindent\textbf{Learning an Approximation:} Recall that $\vec{x}$, the current state of the AV, can be expressed as the position of a moving reference frame attached to the vehicle. \textit{Offline}, a region in front of the AV is sampled, yielding a set of $M$ possible goals $\{\vec{x}_{i}\}_{i=1}^M$, each expressed in relative coordinates.
Then for each goal $\vec{x}_{g,i}$ the reference trajectory connecting them is computed by the original MPC algorithm.
Denote the computed reference trajectory by $\vec{p}_i = [s\ a\ b\ c\ d]^T$ 
Thus we now have a \textit{training} set $\{(\vec{x}_{g,i}, \vec{p}_i)\}_{i=1}^M$.
A neural network $NN_{TP}$ is used to fit the function $x_{goal,i} \mapsto \vec{p}_i$.
\textit{Online}, given an actual target state $\vec{x}_g$ in relative coordinates, the AV computes $NN_{TP}(\vec{x}_g)$ to obtain the parameters of the reference trajectory $\vec{p}_g$ leading to $\vec{x}_g$. Our implementation utilizes a radial basis function network architecture, the benefits of this decision is that the weights can be trained algebraically (via a pseudo-inverse) and each data point is guaranteed to be interpolated exactly. On 145,824 samples in the test set our methodology exhibits a worst-case test error of $0.1\%$ and is capable of generating over 428,000 trajectories per-second.

\subsection{Vehicle-to-Vehicle Communication, Cooperation, and Behavioral Planning}
The development of autonomous vehicles has been propelled by an idealistic notion that the technology can nearly eliminate accidents. The general public expects AVs to exhibit what can best be described as superhuman performance; however, a key component of human driving is the ability to communicate intent via visual, auditory, and motion based cues. Evidence suggests that these communication channels are developed to cope with scenarios in which the fundamental limitations of human senses restrict drivers to \textit{cautious operations} which anticipate dangerous phenomena before they can be identified or sensed.

\noindent\textbf{Components:} In order to carry out V2V communication experiments we augment the F1/10 planning stack with ROS nodes which contain push/pull TCP clients and servers, these nodes extract user defined state and plan information so that it may be transmitted to other vehicles. 

In this research we construct an AV `roundabout' scenario where the center-island obstructs the ego-vehicles view of the other traffic participants. A communication protocol which transmits an object list describing the relative positions of participating vehicles, and a simple indicator function encodes whether given each vehicles preferred next action it is safe to proceed into the roundabout is implemented. Alternative scenarios such as a high-speed merge or highway exit maneuver can also easily be constructed at significantly less cost and risk than real world experiments. The F1/10 platform enables an intermediate step between simulation and real-world testing such that the effects of sensor noise, wireless channel degradation, and actuation error may be studied in the context of new V2V protocols. 

\section{Research: Perception}
\begin{figure*}[t]
    \centering
    \includegraphics[width=0.8\textwidth]{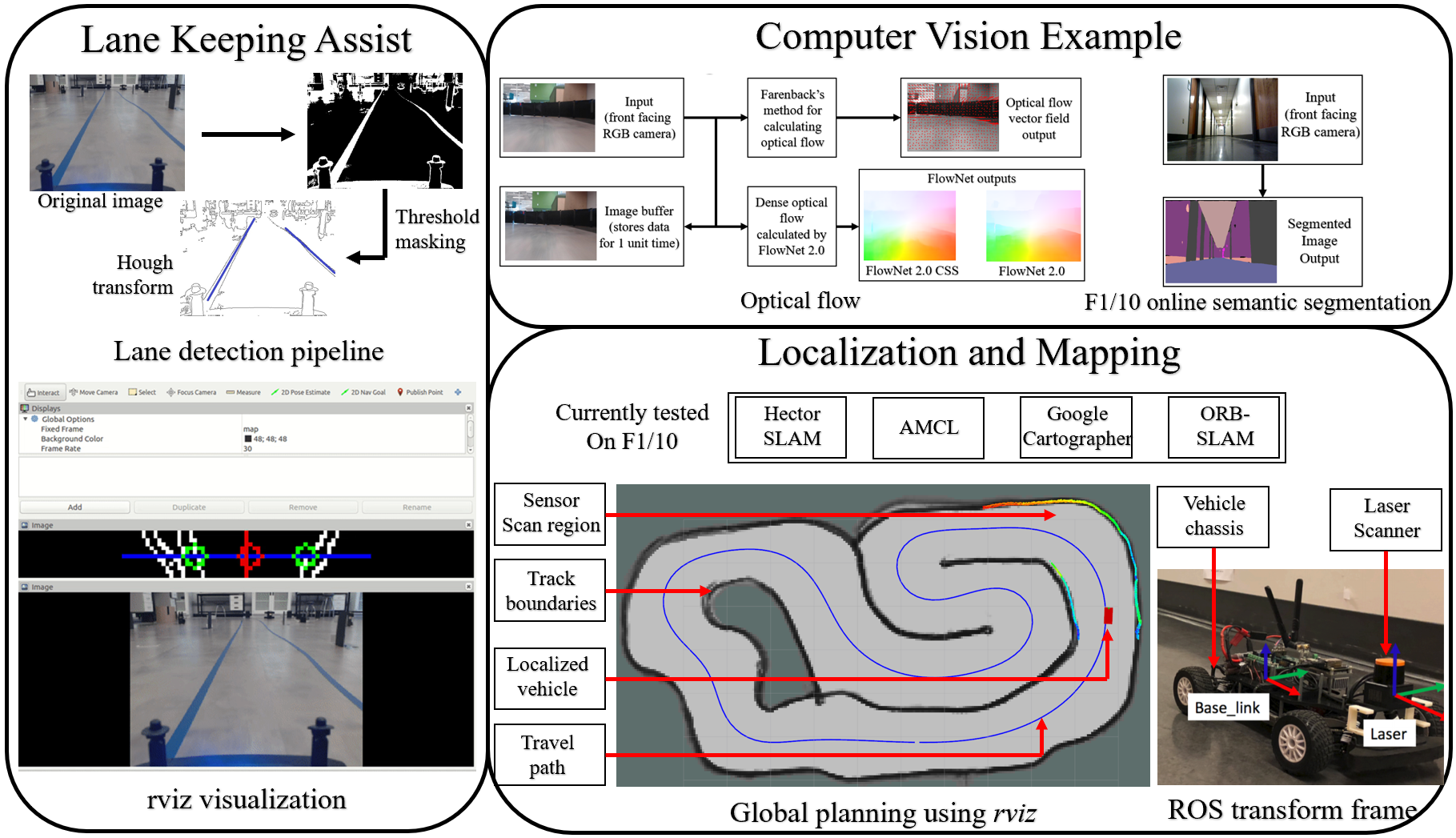}
    \caption{Some perception research enabled by the F1/10 platform (clockwise, starting left); (a) lane following using a monocular camera, (b) optical flow computation using Farenback's method and FlowNet 2.0 and, (c) localization and mapping}
    \label{fig:perception}
\end{figure*}
In this section we highlight how the F1/10 vehicle enables a novel mode of research relative to perception tasks. Although there has been huge progress in low-level vision tasks such as object detection due to effectiveness of deep learning, AVs only perform such tasks in order to enable decisions which lead to safe mobility. In this context the F1/10 vehicle is a unique tool because it allows researchers to measure not just performance of a perception subsystem in isolation, but rather the capabilities of the whole system within its operating regime. Due to the extensive planning and state estimation capabilities already reliably enabled on the car new researchers focused on perception subsystems can enable comparison of a variety of methods on uniform platform in the context of specific driving tasks.

\subsection{Simultaneous Localization and Mapping}
The ability for a robot to create a map of a new environment without knowing its precise location (SLAM) is a primary enabler for the use of the F1/10 platform in a variety of locations and environments. Moreover, although SLAM is a well understood problem it is still challenging to create reliable real-time implementations. In order to allow the vehicle to drive in most indoor environments we provide interface to a state of the art LIDAR-based SLAM package which provides loop-closures, namely Google Cartographer \cite{hess2016real}. Included in our base software distribution are local and global settings which we have observed to work well empirically through many trials in the classroom and at outreach events. In addition we include a description of the robots geometry in an appropriate format which enables plug-and-play operation.
For researchers interested primarily in new approaches to SLAM the F1/10 platform is of interest due to its non-trivial dynamics, modern sensor payload, and the ability to test performance of the algorithm in motion capture spaces (due to the small size of vehicle).

In addition to SLAM packages we also provide an interface to an efficient, parallel localization package which utilizes a GPU implementation of raymarching to simulate the observations of random particles in a known 2D map \cite{walsh17}. The inclusion of this package enables research on driving at the limits of control even without a motion capture system for state estimation. 

\subsection{Computer Vision}
Our distribution of F1/10 software includes the basic ingredients necessary to explore the use of deep learning for computer vision. It includes CUDA enabled versions of PyTorch~\cite{paszke2017pytorch}, Tensorflow~\cite{abadi2016tensorflow}, and Darknet~\cite{redmon2013darknet}. We include example networks for semantic segmentation \cite{DBLP:journals/corr/abs-1803-06815}, object detection \cite{redmon2016you}, and optical flow \cite{ilg2017flownet}; we focus on efficient variants of the state-of-the-art that can run at greater than 10 FPS on the TX2. 
Recently, it has come to light that many DNNs used on vision tasks are susceptible to so called \textit{adversarial examples}, subtle perturbations of a few pixels which to the human eye are meaningless but when processed by a DNN result in gross errors in classification. Recent work has suggested that such adversarial examples are \textit{not} invariant to viewpoint transformations \cite{lu2017no}, and hence \textit{not} a concern. The F1/10 platform can help to enable principled investigations into how errors in DNN vision systems affect vehicle level performance. 

\subsection{Lane keep assist}
The F1/10 platform is designed to work with a wide array of sensors and, among them are USB cameras which enable implementation of lane tracking, and lane keep assist algorithms~\cite{guo2013cadas,satoh2002lane}. 
Utilizing the OpenCV~\cite{bradski2000opencv} libraries. We implemented a lane tracking algorithm~\cite{ruyi2011lane} to run in real-time on the F1/10 on-board computer.
To do so, we created an image processing pipeline to capture, filter, process, and analyze the image stream using the ROS \textit{image\textunderscore transport} package, and designed a ROS node to keep track of the left and right lanes and calculate the geometric center of the lane in the current frame. The F1/10 steering controller was modified to keep track of the lane center using a proportional-derivative-integral (PID) controller. The image pipeline detailed in Fig.~\ref{fig:perception} [Left] is comprised of the following tasks:\\
(a) The raw RGB camera image, in which the lane color was identified by its hue and saturation value, is converted to greyscale and subjected to a color filter designed to set the lane color to white and everything else to black,\\
(b) The masked image from the previous step is sent through a canny edge detector and then through a logical AND mask whose parameters ensured that the resulting image contains only the information about the path,\\
(c) The output from the second step is filtered using a Gaussian filter that reduces noise and is sent through a Hough transformation, resulting in the lane markings contrasting a black background.
The output of the image pipeline contains only the lane markings. 
The lane center is calculated and the F1/10 current heading is compared to the lane center to generate the error in heading. The heading of the car is updated to reflect the new heading generated by the ROS node using a PID controller.


\section{Research: Systems, Simulation, and Verification}
\begin{figure*}[t]
    \centering
    \includegraphics[width=0.8\textwidth]{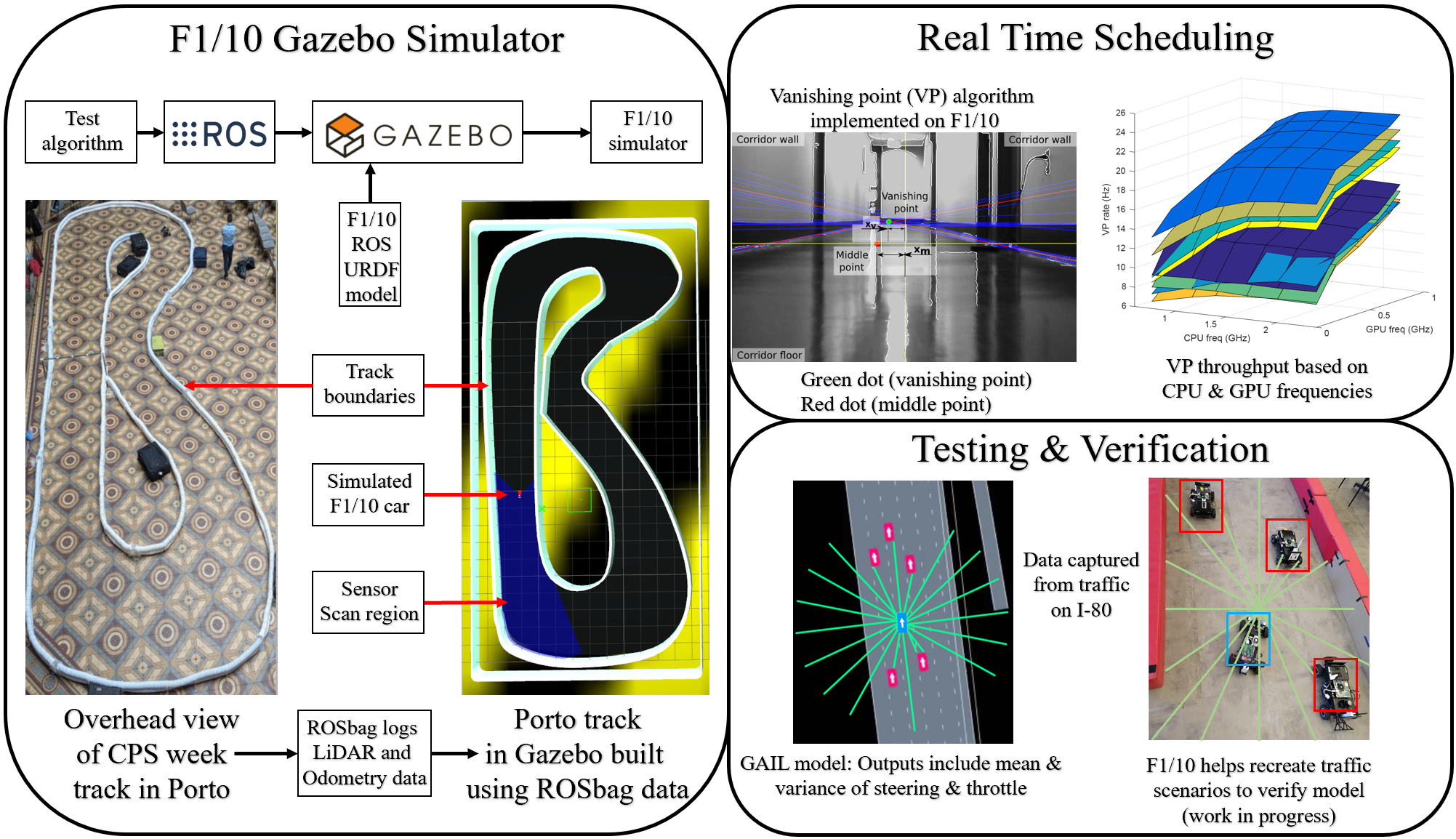}
    \caption{Figure (left) shows an F1/10 car in a simulated environment generated using data from the real world, (right, top) real time scheduling of vanishing point algorithm on the F1/10 onboard computer and, (right, bottom) verifying traffic behavior}
    \label{fig:systems-research}
\end{figure*}
Safety and robustness are key research areas which must make progress in order to deploy commercial AVs. In this section we highlight the tools which we are using to enable simulation, real-time systems research, and verification efforts.



\subsection{Gazebo Racing Simulators}
Why would we want to use a simulator if you have the F1/10 platform? We want to test the car's algorithms in a controlled environment before we bring it into the real world so that we minimize risk of crashing. For instance, if a front steering servo plastic piece were to break, it is necessary to disassemble 20 parts in order to replace it. In fact each of the labs taught in our courses can be completed entirely in simulation first. The added benefit is that researchers and students with resource constraints can still utilize the software stack that we have built. 

We use the ROS Gazebo simulator software \cite{koenig2004design}. From a high level, Gazebo loads a world as a .DAE file and loads the car. Gazebo also includes a physics engine that can determine how the F1/10 car will respond to control inputs, friction forces, and collisions with other obstacles in the environment. The F1/10 simulation package currently provides four tracks, each of which have real world counterparts. 
It is also possible to create custom environments. In the F1/10 reference manual we provide a tutorial on the use of Sketchup to create simple 3D models. More advanced 3D modeling tools such as 3DS Max and Solid Works will also work. Our future work includes a cloud based simulation tool which utilizes the PyBullet \cite{coumans2016pybullet} physics engine and Kubernetes \cite{brewer2015kubernetes} containers for ease of deployment and large scale reinforcement learning experiments.

\subsection{Real-time Systems Research}
Autonomous driving is one of the most challenging engineering problems posed to modern embedded computing systems. It entails processing and interpreting a wide amount of data, in order to make prompt planning decisions and execute them in real-time. Complex perception and planning routines impose a heavy computing workload to the embedded platform, requiring multi-core computing engines and parallel accelerators to satisfy the challenging timing requirements induced by high-speed driving. Inaccuracy in the localization of the vehicles as well as delays in the perception and control loop may significantly affect the stability of the vehicle, and result in intolerable deviations from safe operating conditions. Due to the safety-critical nature of such failures, the F1/10 stack is an ideal platform for testing the effectiveness of new real-time scheduling and task partitioning algorithms which efficiently exploit the heterogeneous parallel engines made available on the vehicle. One example of such research implemented on the F1/10 platform is the AutoV project \cite{xu2017autov}  which explores whether safety critical vehicle control algorithms can be safely run within a virtual environment.

The F1/10 platform also enables real-time systems research which explicitly consider the problem of co-design at the application layer. Specifically the goal is to create planning, perception, and scheduling algorithms which adapt to the context of the vehicle's operating environment. This regime was explored in a study on CPU/GPU resource allocation for camera-based perception and control \cite{pant2015power}. In the experiments performed on the F1/10 platform the objective was to obtain energy-efficient computations for the perception and estimation algorithms used in autonomous systems by manipulating the clock of each CPU core and the portion of the computation which would be offloaded to the a GPU. 
These knobs allow us to leverage a trade-off between computation time, power consumption and output quality of the perception and estimation algorithms. 
In this experiment, a vanishing point algorithm is utilized to navigate a corridor. 
The computation is decomposed into three sequential components, and we study how its runtime and power consumption are affected by whether each component is run on a GPU or CPU, and the frequency at which it is executed.
Results highlight CPU/GPU allocation and execution frequencies which achieve either better throughput or lower energy consumption without sacrificing control performance.
The possible set of operating points and their effect on the update rate and power consumption for the vanishing point algorithm are shown in Fig.~\ref{fig:systems-research} [Middle]. 


\subsection{Monitoring, Testing, \& Verification}

F1/10 can be used to support and demonstrate advances in formal verification and runtime monitoring. 
\\
\textbf{Real-time verification}.
Rechability analysis is a technique for rigorously bounding a system's future state evolution, given that its current state $x(t)$ is known to be in some set $X(t)$.
The uncertainty about the system's current state is due to measurement noise and actuation imperfections.
Being able to ascertain, rigorously, bounds on the system state over $[t,t+T]$ despite current uncertainty allows the car to avoid unsafe plans.
Calculating the system's \textit{reach set}, however, can be computationally expensive and various techniques are proposed to deal with this issue, but very few have been explicitly aimed at real-time operation, or tested in a real-life situation.
The F1/10 platform enables such testing of reachability software in a real-world setup, with the code running along with other loads on the target hardware.
\\
\textbf{Runtime monitoring}
Good design practice requires the creation of \textit{runtime monitors}, which are software functions that monitor key properties of the system's behavior in real-time, report any violations, and possibly enforce fail-safe behavior.
Increased sophistication in the perception and control pipelines necessitates the monitoring of complex requirements, which range from enforcing safety and security properties to pattern matching over sensor readings to help perception~\cite{Abbas18Emsoft}.
A promising direction is to generate these complex monitors automatically from their high-level specification~\cite{havelundruntime,bartocci2018specification,rewriting-techniques,ulus2018sequential,havelund2002synthesizing,basin2018algorithms,dokhanchi2014line}.
These approaches have been implemented in standalone tools such as \cite{montre,mop-overview,basin2011monpoly,reger2015marq,Anna10staliro}.
For robotic applications, it will be necessary to develop a framework that handles specifications in a unified manner and generates efficient monitoring ROS nodes to be deployed quickly in robotic applications.
Steps in this direction appear in ROSMOP\footnote{https://github.com/Formal-Systems-Laboratory/rosmop}~\cite{mop-overview}, and in REELAY\footnote{https://github.com/doganulus/reelay}.
The F1/10 platform is ideal for testing the generated monitors' efficiency. 
Its hardware architecture could guide low-level details of code generation and deployment over several processors.
The distributed nature of ROS also raises questions in distributed monitoring.
Finally, F1/10 competitions could be a proving ground for ease-of-use: based on practice laps, new conditions need to be monitored and the corresponding code needs to be created and deployed quickly before the next round. 
This would be the ultimate test of user-friendliness.

\noindent\textbf{Generating Adversarial Traffic}
Because F1/10 cars are reduced-scale, cheaper and safer to operate than full-scale cars, they are a good option for testing new algorithms in traffic, where the F1/10 cars provide the traffic.
E.g. if one has learned a dynamic model of traffic in a given area, as done in \cite{okelly2018} then that same model can drive a fleet of F1/10 cars, thus providing a convincing setup for testing new navigation algorithms.
This fleet of cars can also allow researchers to evaluate statistical claims of safety, since it can generate more data, cheaply, than a full-scale fleet.

\section{From F1/10 to full-scale AVs}
\begin{figure}[t]
    \centering
    \includegraphics[width=0.8\columnwidth]{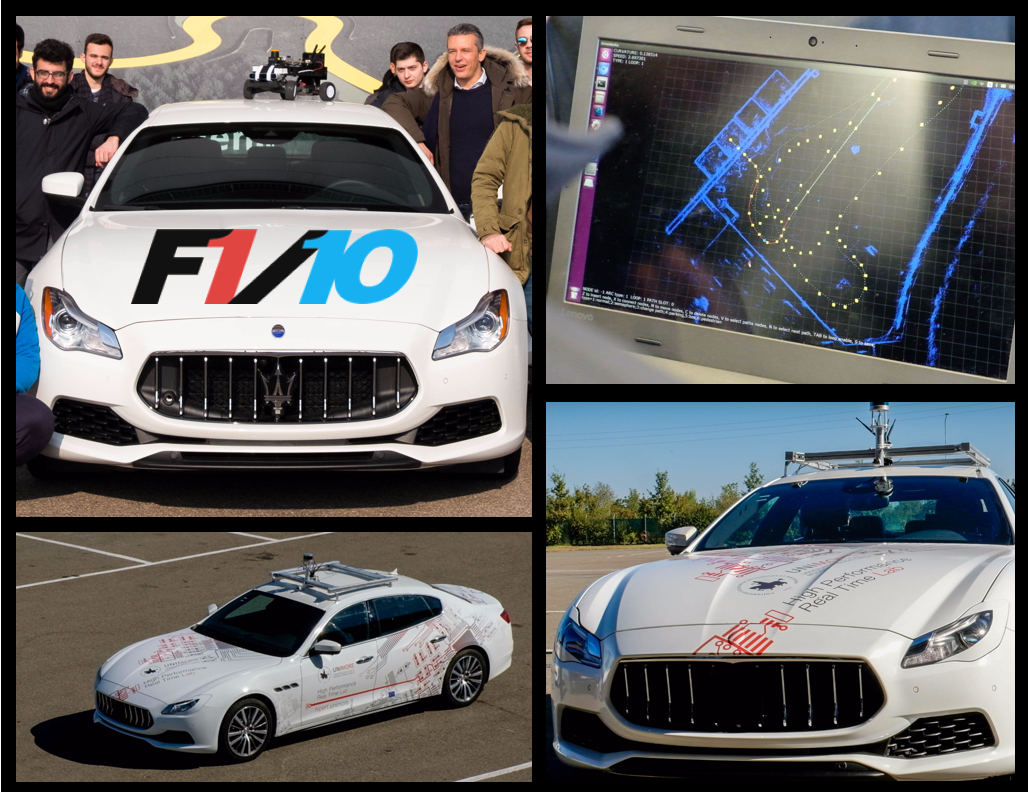}
    \caption{Open source perception, planning and control pipeline of F1/10 platform has been successfully applied in the design of full-scale autonomous racecars at UNIMORE, Italy}
    \label{fig:real-car}
\end{figure}
The open source AV stack provided by the F1/10 project represents 
an excellent starting point to implement the perception-planning-actuation pipeline of a full scale vehicle. 
Fig.\ref{fig:real-car} shows a vehicle prototype realized by the HiPeRT Lab of the University of Modena which extends the F1/10 stack with the required drivers and routines to process data from six (Sekonix) cameras, a 3D Lidar (Velodyne VLP-16) and a differential GPS receiver. 
The primary controller is based on NVIDIA's Drive PX Autocruise platform, the automotive-grade version of the Jetson TX2 board adopted in the F1/10 project. 
The car is able to automatically exit a parking lot, navigate autonomously in roundabouts and line-marked paths while avoiding detected obstacles, stop at traffic signals when required, and park itself in a user-defined slot. The first version of the HiPeRT autopilot utilzes a pure pursuit trajectory tracker developed within the F1/10 framework, and a model predictive controller for trajectory generation. The Gazebo simulator (and an alternative version based on Unity) have been adopted to test the HiPeRT autopilot before deploying it to the real car controller. The initial prototype, like F1/10, includes modules such as lane keep assist, DNNs for object detection, a basic obstacle avoidance planner, SLAM algorithms (HectorSLAM, Cartographer, GraphSLAM, particle filters, etc.), Camera/Lidar sensor fusion, and V2I communication support. In the current version of the car the ROS-based meta operating system is replaced by a stack with better real-time guarantees. Nevertheless, the routines made available by the F1/10 project were instrumental in enabling the deployment of a working AV prototype in a very limited time. Additionally the F1/10 platform proved to be an ideal sandbox in which to incrementally iterate on the development of the various components.

\section{F1/10 Education and Compeitions}
\begin{figure*}[t]
    \centering
    \includegraphics[width=0.85\textwidth]{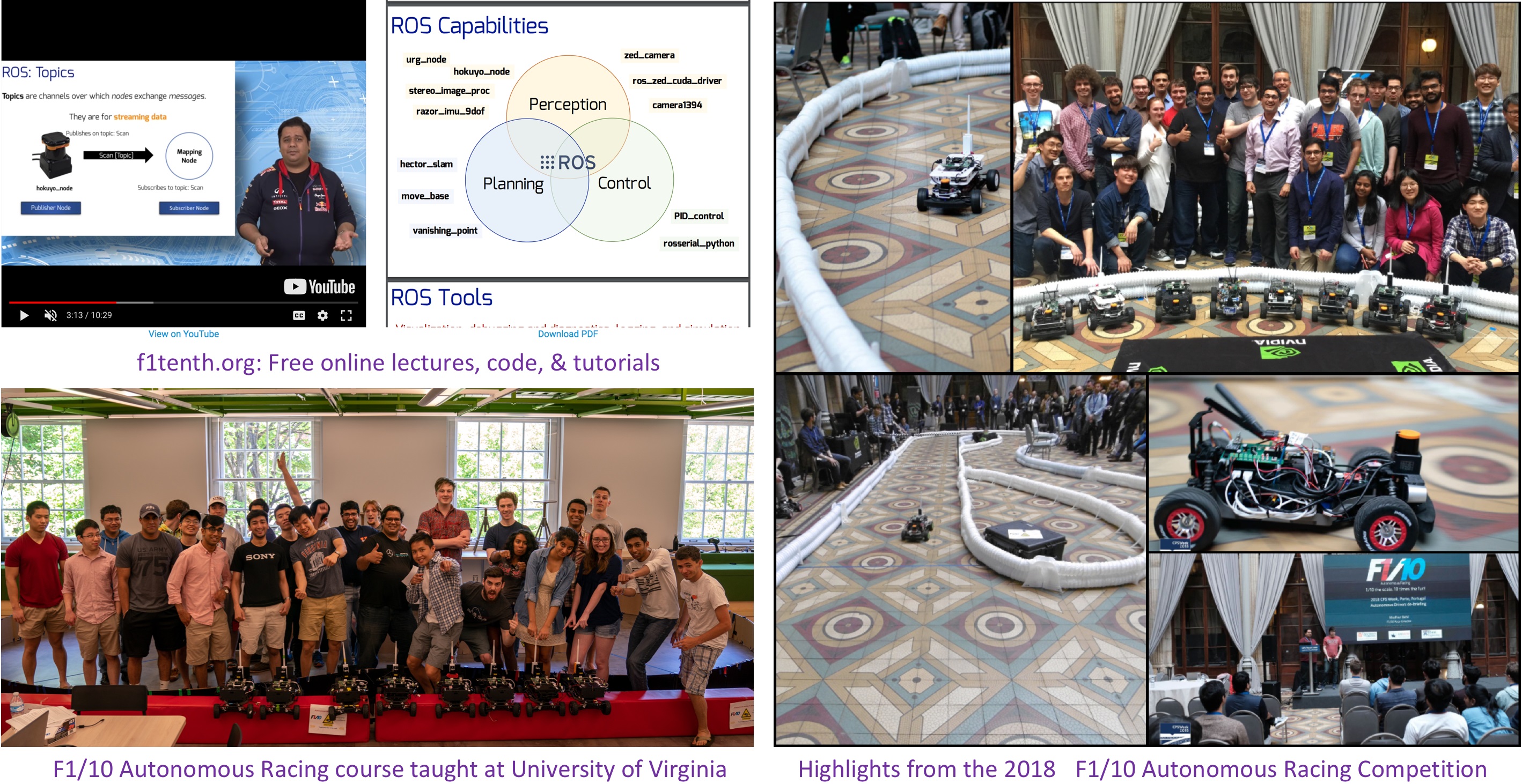}
    \caption{The F1/10 testbed instrument is enabling K-12, undergrad, and graduate outreach through our online courses and MOOCs, autonomous racing competitions, summer schools, and hackathaons.}
    \label{fig:outreach}
\end{figure*}

The F1/10 platform is the basis of a full-semester class on autonomous driving at the University of Virginia, University of Pennsylvania and at Oregon State University.
In addition, our open source course material has been taught at UT Austin (2016) and Clemson University (2017).
The online videos from UVA have been viewed thousands of times since launch.
The courses are offered to graduate and advanced undergraduate students with backgrounds in any of the following: electrical engineering, mechanical engineering, robotics, embedded systems, and computer science.
Our goal is to present an example of a class which teaches the entire stack for autonomous driving: from assembling the electrical components and sensors, to programming the car at two levels of complexity.
%
Students who enroll in these courses will learn technologies that drive today's research AVs, and have the confidence to develop more sophisticated approaches to these problems on their own.
Importantly, the students become familiar with the system \textit{as a whole}, and encounter integration problems due to non-real-time performance, mechanical limitations, and sensor choices - this is why the students are divided into inter-disciplinary teams. 

The class is project-based, with the formal lectures introducing the techniques that students will code and implement on the car. 
In the first half of the semester, the students are guided to the point where their car can navigate an environment with static obstacles. 
In the first week they build the car and can control it manually. 
Then they successively tackle LiDAR data processing with gap-finding, 
coordinate transformations, 
reference tracking, 
Electronic Speed Control,
localization with scan matching, 
Simultaneous Localization and Mapping, 
and path planning.
We also tackle the thorny question of moral decision-making for autonomous systems.
For this, we assign readings from the humanities and sciences on topics like responsibility and moral agency, and have in-class guided discussions on the possibility of ethics for autonomous robots.
The course culminates in a F1/10 `battle of algorithms' race among the teams.


\vspace{-12pt}
\subsection{The F1/10 Competition}
Few things focus the mind and excite the spirit like a competition. 
In the early days of racing, competitors first had to build their vehicles before they could race them. It was thus as much an engineering as a racing competition. We want to rekindle that competitive spirit.

For the past three years, we have been organizing the F1/10 International Autonomous Racing Competition, the first ever event of its kind. 
The inaugural race was held at the 2016 ES-Week in Pittsburgh, USA; followed by another race held during Cyber-Physical Systems (CPS) Week in April 2018, in Porto, Portugal. The third race was held at the 2018 ES-Week in October in Turin, Italy,  Figure~\ref{fig:outreach} [Right]). 
Every team builds the same baseline car, following the specifications on \url{f1tenth.org}.
From there, they have the freedom to deploy any algorithms they want to complete a loop around the track in the fastest time, and to complete the biggest number of laps in a fixed duration. 
Future editions of the race will feature car-vs-car racing.

So far, teams from more than 12 universities have participated in the F1/10 competition, including teams from KAIST (Korea), KTH (Sweden), Czech Technical University, University of Connecticut, Seoul National University, University of New Mexico, Warsaw university of Technology, ETH Zurich, Arizona State University, and Daftcode (a Polish venture building company). 

\section{Conclusion and Discussion}
The paper presents a new open-source and widely used 1/10 scale autonomous vehicle testbed called F1/10. 
All the instructions to build, drive, and race the F1/10 car are freely available on \url{f1tenth.org}. 
F1/10 uses a modular hardware and software design enabling researches to shape and use the platform to fit their needs. 
The default configuration houses several sensors and a powerful on-board GPU - similar to a full scale car. 
The chassis of the F1/10 platform provides realistic dynamics so that researchers can test their algorithms on the 1/10 scale safely and cost-effectively. The open-source ROS based software stack makes it very easy for beginners to get up to speed with autonomous driving behavior and build on existing capabilities. We show three representative examples of the kind of research that is easily enabled by the F1/10 platform - obstacle avoidance, land keep assist, and end-to-end autonomous driving. 
F1/10 has slowly become a popular instrument to make autonomy accessible and bring it to the classroom. Dozens of research groups have built their cars using instructions and videos available on the F1/10 web page. We also present highlights from the International F1/10 Autonomous Racing Competitions, which have been previously held at prominent CPS and Embedded Systems venues. 
The F1/10 autonomous platform is the building block and the vehicle for educating tomorrow's engineers on the interlocking concerns of performance, control, and safety for autonomous systems, and in particular for autonomous vehicles.
It will also be the meeting point where a community of researchers from different backgrounds can develop their ideas in a way that emphasizes prototyping and real-world testing.  As the community continues to grow,  so does the range of possibilities of what we can discover and create.


\bibliographystyle{acm}
\bibliography{ms}

\end{document}